\documentclass[runningheads]{llncs}
\usepackage{floatrow}
\newfloatcommand{capbtabbox}{table}[][\FBwidth]
\usepackage{pifont}
\newcommand{\cmark}{\ding{51}}
\newcommand{\xmark}{\ding{55}}
\usepackage{graphicx}
\usepackage{amsmath,amssymb} 
\usepackage{color}
\usepackage[width=122mm,left=12mm,paperwidth=146mm,height=193mm,top=12mm,paperheight=217mm]{geometry}

\usepackage{subfigure}

\begin{document}
\pagestyle{headings}
\mainmatter
\def\ECCV18SubNumber{808} 

\title{
Learning Beyond Human Expertise with Generative Models for Dental Restorations
} 

\titlerunning{ }

\authorrunning{ }

\author{Jyh-Jing Hwang$^1$ \and Sergei Azernikov$^2$ \and Alexei A. Efros$^1$ \and Stella X. Yu$^1$}
\institute{University of California, Berkeley \and Glidewell Dental Labs}

\maketitle

\begin{abstract}

Computer vision has advanced significantly that many discriminative approaches such as object recognition are now widely used in real applications.  We present another exciting development that utilizes generative models for the mass customization of medical products such as dental crowns.  
In the dental industry, it takes a technician years of training to design synthetic crowns that restore the function and integrity of missing teeth.  Each crown must be customized to individual patients, and it requires human expertise in a time-consuming and labor-intensive process,  even with computer assisted design software.
We develop a fully automatic approach that learns not only from human designs of dental crowns, but also from natural spatial profiles between opposing teeth.  The latter is hard to account for by technicians but important for proper biting and chewing functions.  
Built upon a Generative Adversarial Network architecture (GAN), our deep learning model predicts the customized crown-filled depth scan from the crown-missing depth scan and opposing depth scan. We propose to incorporate additional space constraints and statistical compatibility into learning.
Our automatic designs exceed human technicians' standards for good morphology and functionality, and our algorithm is being tested for production use.

\end{abstract}
\section{Introduction}
\label{sec:intro}

Computer vision has advanced so significantly that many discriminative approaches such as object detection, object recognition, and semantic segmentation, are now successfully deployed in real applications~\cite{krizhevsky2012imagenet,he2016deep}.  Generative approaches, focusing on generating photo-realistic images, have largely remained as research topics.  We present a first major success of generative models applied in the field of mass customization of medical products, specifically, for dental restoration~\cite{summitt2006fundamentals}.

\begin{figure}[t]
    \centering
    \includegraphics[width=0.9\linewidth]{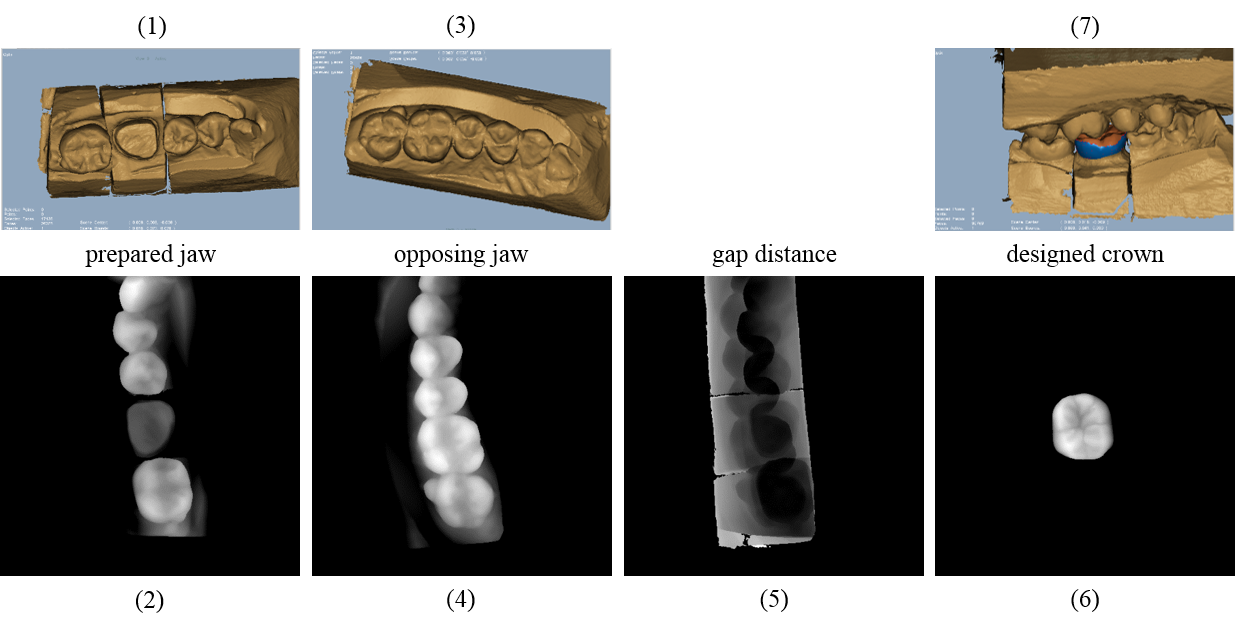}
    \caption{Illustration of dental crown design stages. 3D models of the prepared jaw, opposing jaw, and the crown-filled restoration are shown at the top, and their 2D depth images from a particular plane shown below.  The gap distances are computed from opposing jaws.}
    \label{fig:procedure}
\end{figure}

In dental restoration, the dentist first prepares the patient's teeth by removing the decayed portion of the dentition. He then takes an impression of the prepared tooth and its surrounding structures, either physically or digitally by using an intra-oral 3D scanner. The captured data is used to produce a full crown or inlay.  Dental restoration has to satisfy a number of requirements~\cite{padbury2003interactions} which are important for successful outcomes:
\begin{enumerate}
    \item It must perfectly fit patient's dentition;
    \item It has to provide chewing functionality;
    \item It should have an aesthetically plausible shape.
\end{enumerate}
Computer aided design (CAD)~\cite{zheng2011novel} technologies introduced in the last decade into dental industry have significantly facilitated achievements of these requirements. However, human assistance is still much required in the current process. Dental CAD is usually based on a predefined template library of ideal tooth models. The template model is positioned on the prepared site and adjusted to the patient's anatomy. In the dental restoration procedure illustrated in Fig.~\ref{fig:procedure}, the designer needs to evaluate the crown and make adjustments manually.

In order to build an automatic dental CAD system, human expertise needs to be integrated into the software. One approach is to build a comprehensive set of rules that would include all the nuances known to the experienced dental professionals and formulate it in a way that machines can understand. This is a very tedious task, and obviously feasible only when this set of rules can be provided. A different approach is to build a system capable of learning from a large number of examples without explicit formulation of the rules. 


We follow the latter data-driven deep learning approach and formulate the dental restoration task as a conditional image prediction problem. We represent the 3D scan as a 2D depth image from a given plane.  The prepared crown-missing depth image serves as the input condition, and the technician-designed crown-filled depth image serves as the ground-truth output prediction.    That is, we can learn from the technicians and capture the human design expertise with a deep net that translates one image into another~\cite{pix2pix2017}.

However, there are a few challenges to which technicians have no good solutions aside from trials and errors, e.g. how to design natural grooves on the tooth surface, and how to make a few points of contact with the opposing teeth to support proper biting and chewing~\cite{groten2000determination}.

The exciting aspect of our work is that we can learn beyond human expertise on dental restoration by learning natural fitting constraints from big data.  
We propose to incorporate both hard and soft functionality constraints for ideal dental crowns: The former captures the physical feasibility where no penetration of the crown into the opposing jaw is allowed, and the latter captures the natural spatial gap statistics between opposing jaws where certain sparse contacts are desired for proper biting and chewing.

We accomplish the conditional image prediction task using a Generative Adversarial Network (GAN) model~\cite{gan2014,cgan2014,pix2pix2017,context2016} with novel learning losses that  enforce the functionality constraints that were beyond the reach of human experts.  We compare our automatic predictions with technicians' designs, and evaluate successfully on a few metrics of interest to practitioners in the field.  We pass the ultimate field test and our algorithm is currently being tested for production .

\section{Related Work}
\label{sec:work}

\textbf{Generative models.}
Modeling the natural image distribution draws many interests in computer vision research. Various methods have been proposed to tackle this problem, such as restricted Boltzmann machines~\cite{smolensky1986information}, autoencoders~\cite{hinton2006reducing,vincent2008extracting}, autoregressive models~\cite{efros1999texture}, and generative adversarial networks~\cite{gan2014}. Variational autoencoders~\cite{kingma2013auto} capture the stochasticity of the distribution by training with reparametrization of a latent distribution. Autoregressive models~\cite{efros1999texture,oord2016pixel,van2016conditional} are effective but slow during inference as they generate an image pixel-by-pixel sequentially. Generative adversarial networks (GANs)~\cite{gan2014}, on the other hand, generate an image by a single feedforward pass with inputs of random values sampled from a low-dimensional distribution. GANs introduce a discriminator, whose job is to distinguish real samples from fake samples generated by a generator, to learn the natural image distribution. Recently, GANs have seen major successes~\cite{arjovsky2017towards,chen2016infogan,denton2015deep,donahue2016adversarial,dumoulin2016adversarially,radford2015unsupervised,reed2016generative,zhang2017stackgan,zhao2016energy,zhu2016generative,karras2017progressive} in this task. \\

\noindent \textbf{Conditional image generation.} All of the methods mentioned above can be easily conditioned. For example, conditional VAEs~\cite{sohn2015learning} and autoregressive models~\cite{oord2016pixel,van2016conditional} have shown promising results~\cite{guadarrama2017pixcolor,walker2016uncertain,xue2016visual}.  Prior works have conditioned GANs~\cite{cgan2014} to tackle various image-to-image generation problems~\cite{wang2016generative,mathieu2015deep,yoo2016pixel,ledig2016photo}. Among them, image-to-image
conditional GANs (pix2pix)~\cite{pix2pix2017,zhu2017unpaired,zhu2017toward} have led to a substantial boost in results under the setting of spatial correspondences between input and output pairs.
\\

\noindent \textbf{Statistical features.}  Statistical features can be traced back to hand-crafted features, including Bag-of-Words~\cite{lazebnik2006beyond}, Fisher vector~\cite{perronnin2010improving}, Second-order pooling~\cite{carreira2012semantic}, etc. Such global context features compensate for hand-crafted low-level features. Along this line of research, there are a number of deep learning methods that tried to incorporate statistical features into deep neural networks. For example, the deep Fisher network~\cite{simonyan2013deep} incorporates Fisher vector and orderless pooling~\cite{gong2014multi} combines with Vector of Locally Aggregated Descriptors. Both methods simply treat features by deep networks as off-the-shelf features. More recently, several works \cite{wang2016learnable,ustinova2016learning} have shown successes in incorporating histogram learning into deep neural networks for classification tasks and feature embedding.

\section{Our Approach}
\label{sec_method}

\begin{figure}[t]
    \centering
    \includegraphics[width=1.0\textwidth]{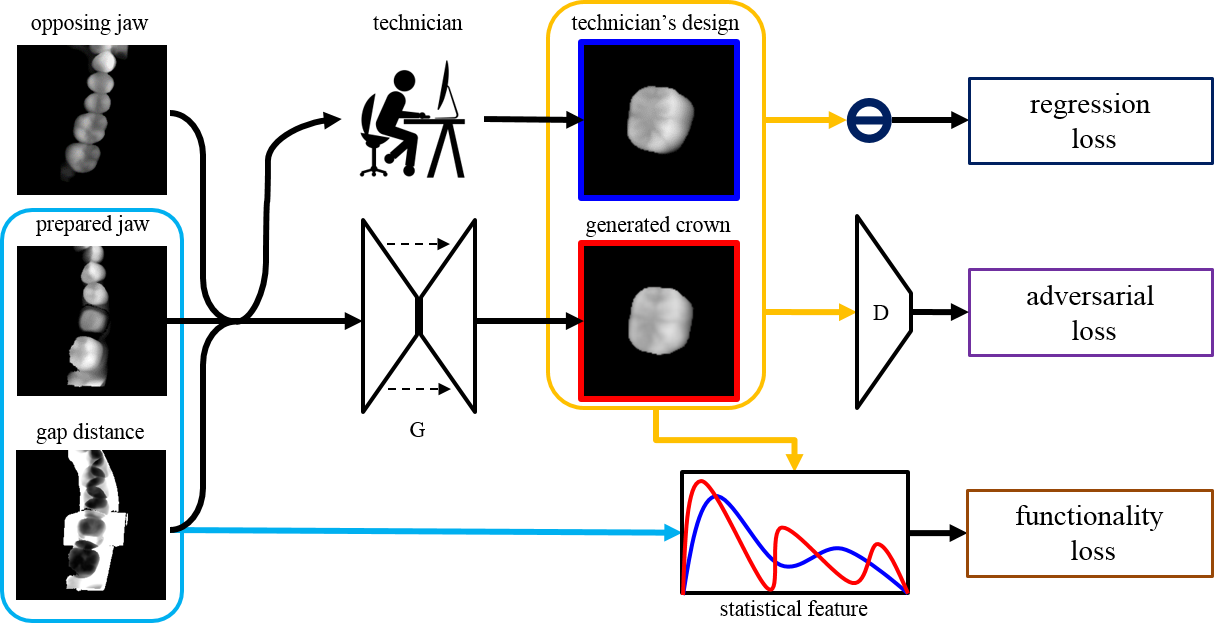}
    \caption{Diagram illustration of the proposed model. We propose a functionality loss with space information (in the blue box) to learn the functionality of technician's designs. Please refer to Fig.~\ref{fig:histogram} for the computation diagram of the functionality loss. (For all the crowns, we zoom in the images by a factor of $2$ for better visualization.)}
    \label{fig:method}
\end{figure}

Our dental crown design pipeline as shown in Fig.~\ref{fig:procedure} is as follows. We first create 2D scan images of the prepared jaw, opposing jaw, and the gap distances between two jaws from the original intra-oral 3D scan model. We then apply the proposed generative model as shown in Fig.~\ref{fig:method} to predict a best fitting crown, which is learned from good technician's design. We then transform the generated 2D crown surface back to 3D model using CAD tools. If a generated 3D crown passes all the spatial constraints, it is ready to be produced.

If we directly apply 2D image generation models such as pix2pix model~\cite{pix2pix2017}, the generated crown does fit into neighboring teeth yet hardly satisfies the full functionality of a tooth. The ideal crowns should not penetrate into the opposing teeth in the 3D model while maintaining few contact points to tear and crush food. Hence, we propose to incorporate a functionality loss to tackle this problem.

We organize this section as follows. We briefly review the conditional generative adversarial model, or specifically the pix2pix model~\cite{pix2pix2017} in section~\ref{sec:pix2pix}. Then in section~\ref{sec:conditioning}, we condition the model with space information as the first attempt to tackle the problem. Finally, we formulate the functionality constraints and introduce the functionality loss using statistical features with few variants in section~\ref{sec:functionality}. The proposed model is summarized in Fig.~\ref{fig:method}.

\subsection{Conditional Generative Adversarial Network}
\label{sec:pix2pix}

The recently proposed pix2pix model~\cite{pix2pix2017} has shown promising results in the image-to-image
translation setting in various tasks. The idea is to leverage conditional generative adversarial networks~\cite{cgan2014} to help refine the generator so as to produce a perceptually realistic result. 
The conditional adversarial loss with an input image $x$ and ground truth $y$ is formulated as
\begin{equation}
    \mathcal{L}_{cGAN}(G, D) = \mathbb{E}_{x,y}[\log D(x,y)] + \mathbb{E}_{x,z}[\log (1-D(G(x,z))],
\end{equation}
where $G$ attempts to minimize this loss against $D$ that attempts to maximize it, i.e., $G^*=\arg\min_G\max_D \mathcal{L}_{cGAN}(G,D)$.

The adversarial loss encourages the distribution of the generated samples to be close to the real one. However, the loss does not penalize directly the instance-to-instance mapping. The L1 regression loss is thus introduced to ensure the generator to learn the instance-to-instance mapping, which is formulated as
\begin{equation}
    \mathcal{L}_{L1}(G)= \mathbb{E}_{x,y,z}[|| y - G(x,z) ||_1].
\end{equation}

The final loss function combines the adversarial loss and L1 regression loss, balanced by $\lambda_{L1}$:
\begin{equation}
\label{eq:objective}
    G^*=\arg\min_G\max_D \mathcal{L}_{cGAN}(G,D) + \lambda_{L1} \mathcal{L}_{L1}(G).
\end{equation}

\subsection{Conditioned on Space Information}
\label{sec:conditioning}

Fitting the crown with neighboring teeth does not satisfy the full functionality of a tooth other than the appearance and implantation. Also, we have to consider how the crown interacts with opposing teeth to chew and bite. That is, to generate a well functioning crown, the information of corresponding opposing teeth and per-pixel gap distances between two jaws is needed. 

One straightforward way to feed this information is through the formulation of conditional variables. Besides conditioned on prepared jaws, the network can also be conditioned on space information such as the opposing jaw $\tilde{x}$ and gap distances $d$ between two jaws. We can then re-formulate the conditional adversarial loss as
\begin{equation}
    \mathcal{L}_{cGAN}(G, D) = \mathbb{E}_{x,\tilde{x},d,y}[\log D(x,\tilde{x},d,y)] + \mathbb{E}_{x,\tilde{x},d,z}[\log (1-D(G(x,\tilde{x},d,z))].
\end{equation}
Also, the L1 reconstruction loss is reformulated as
\begin{equation}
    \mathcal{L}_{L1}(G)= \mathbb{E}_{x,\tilde{x},d,z,y}[|| y - G(x,\tilde{x},d,z) ||_1],
\end{equation}
and the final objective remains the same as of Equation~\ref{eq:objective}.


\subsection{Functionality-Aware Image-to-Image Translation}
\label{sec:functionality}

\begin{figure}[t]
    \centering
    \includegraphics[width=0.9\textwidth]{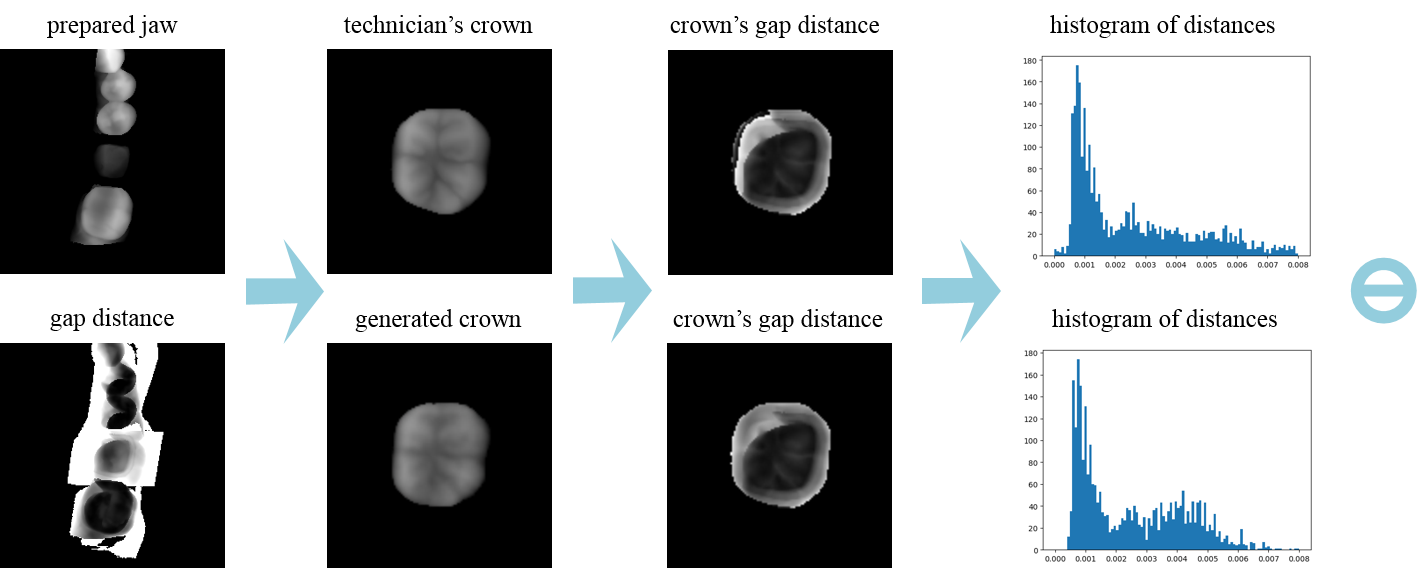}
    \caption{Computation diagram for computing the histogram loss. We calculate the distance map in crown's region and the corresponding histogram with piece-wise differentiable kernels. Finally we compute the $\chi^2$ distance between two histograms. (The unit in x-axis of histograms is meter.)}
    \label{fig:histogram}
\end{figure}

Conditioning on space information does not make the constraints explicit, i.e., the network will fit generated crowns into plausible space yet is unaware of the actual critical constraints. To formulate the functionality constraints, we have to reconstruct gap distances with the generated crown $\hat{y}$, which can be calculated given the prepared jaw and input gap distances:
\begin{equation}
    f(d,x,\hat{y}) = d + \gamma(\hat{y}-x) \text{ where  } \hat{y} > 0,
\end{equation}
where $\gamma$ is a scaling parameter that maps pixel values to real distance values (in millimeters), or $3.14\times 10^{-2}$ in our case.

Now we discuss how to model the constraints. On one hand, the ideal crown should not touch the opposing teeth, or with reconstructed gap distance less than $0$. Otherwise, the crown will penetrate into the opposing teeth when transformed back to 3D models. This crown is then considered over-grown and will hurt the interactions of neighboring teeth. On the other hand, the ideal crown should maintain few contact points so as to tear and crush food. In other words, the reconstructed distance map should follow certain distribution of critical areas. We can thus model the two functionality constraints as
\begin{equation}
    \begin{cases}
    f(d,x,G) > 0, \\
    \hat{f}(d,x,G) \approx \hat{f}(d,x,y),
    \end{cases}
\end{equation}
where $\hat{f}$ denotes critical regions. The critical regions are defined as those pixels with gap distance values less than or equal to minimal $5\%$ of $f$'s overall distance values. This ratio is chosen given the real critical distance in practice, which is $0.5$ millimeters.

To incorporate the functionality constraints into learning, we consider matching the distribution of the gap distances. The reasons have two folds. Firstly, the reconstructed distance map does not need to match the ground truth distance map pixel-by-pixel because this mapping has already been modeled using L1 regression loss. Secondly, to satisfy the constraints, modeling specific locations of contact points are not necessary. By relaxing the spatial correspondences, the model is allowed to explore in larger optimization space. Therefore, we propose a histogram loss, formulated with $\chi^2$ distance, to model the functionality as
\begin{equation}
    \mathcal{L}_{H}(G)= \mathbb{E}_{x,\tilde{x},d,z,y}\left[\sum_i\frac{\big(h_i(f(d,x,G)) - h_i(f(d,x,y))\big)^2}{\max\{1,h_i(f(d,x,y))\}} \right],
\end{equation}
where the $i$-th bin is modeled by a differentiable piece-wise linear function $h_i(d)$:
\begin{equation}
    h_i(d) = \text{sum} \big \{ \max\{0, 1-|d-c_i| \times l_i\} \big \},
\end{equation}
where $c_i$ and $l_i$ are the center and width of the $i$-th bin, respectively. This histogram loss will then back-propagate errors with slope $\pm l_i$ for any pixel in $d$ that falls into the interval of $(c_i-l_i, c_i+l_i)$. The computation diagram is illustrated in Fig.~\ref{fig:histogram}.

The final objective is therefore
\begin{equation}
    G*=\arg\min_G\max_D \mathcal{L}_{cGAN}(G,D) + \lambda_{L1} \mathcal{L}_{L1}(G) + \lambda_H \mathcal{L}_{H}(G),
\end{equation}
where the histogram loss is balanced by $\lambda_H$. \\

Since the functionality is related only to critical regions, it is beneficial to reinforce the histogram loss on certain ranges (preferably on the critical regions), or apply weighting $w_i$ on the $i$-th bin as
\begin{equation}
    \mathcal{L}_{\hat{H}}(G)= \mathbb{E}_{x,\tilde{x},d,z,y}\left[\sum_i w_i\frac{\big(h_i(f(d,x,G)) - h_i(f(d,x,y))\big)^2}{\max\{h_i(f(d,x,y)),1\}} \right],
\end{equation}
The weighting should be chosen by analyzing gap distances and considering critical distances in practice. The details are described in the experimental section.

One property of the proposed histogram loss is spatial invariance. That is, the crown surface is allowed to change dramatically. Sometimes this property is undesirable because it might produce unnecessary spikes on the surface. To make the spatial invariance less and the surface smoother, we propose to incorporate second-order information into the histogram loss. We formulate the second-order operation as local averaging. So the second-order histogram loss is defined as
\begin{equation}
    \mathcal{L}_{H^2}(G)= \mathbb{E}_{x,\tilde{x},d,z,y}\left[w_i\frac{\big(h_i(\bar{f}(d,x,G)) - h_i(\bar{f}(d,x,y))\big)^2}{\max\{h_i(\bar{f}(d,x,y)),1\}} \right],
\end{equation}
where $\bar{f}(d,x,y)$ denotes $f(d,x,y)$ after average pooling.

\section{Experiment}
\label{sec:exp}

We conduct experiments to evaluate and verify our proposed approach. We describe the dataset, network architecture, and training procedure in section~\ref{sec:implementation} and the experimental settings in section~\ref{sec:analysis}. We assess the quality of generated crowns in section~\ref{sec:quality} using ideal technicians' designs as ground truth. We then use the hard cases, where technicians' designs fail, to evaluate our approach and show that our approach greatly reduces failure rates in section~\ref{sec:penetration} and improves functionality in section~\ref{sec:biting}.

\subsection{Implementation Detail}
\label{sec:implementation}

\textbf{Dataset:}
Our dental crown dataset contains 1500 training, 1570 validation, and 243 hard testing cases. For every case, there are a scanned prepared jaw, a scanned opposing jaw, a gap distance map between two jaws, and a manually designed crown (treated as ground truth for training and validation sets). The missing tooth is the number $36$ in the International Standards Organization Designation System, but other teeth can be modeled using the same method. All the hyper-parameters are chosen given the validation results. The testing cases, albeit not as many cases as in the other sets, are especially hard because the manually designed crowns fail the penetration test. We demonstrate the effectiveness of our proposed method mainly on the hard testing cases. \\

\noindent \textbf{Network Architecture:} 
We follow closely the architecture design of pix2pix~\cite{pix2pix2017}. For generator $G$, we use the U-Net~\cite{unet2015} architecture, which contains an encoder-decoder
architecture, with symmetric skip connections. It has been shown to produce strong results when there is a spatial correspondence between input and output pairs. The encoder architecture is:

\texttt{C64-C128-C256-C512-C512-C512-C512-C512}, where \texttt{C\#} denotes a convolution layer followed by the number of filters.
The decoder architecture is:

\texttt{CD512-CD512-CD512-C512-C256-C128-C64}, where \texttt{CD\#} denotes a deconvolution layer followed by the number of filters. After the last layer in the decoder, a convolution is applied to map to the number of output channels, followed by a Tanh function. BatchNorm is applied after every convolutional layer except for the first C64 layer in the encoder. All ReLUs in the encoder are leaky with slope $0.2$, while ReLUs in the decoder are regular.

For discriminator $D$, the architecture is:

\texttt{C64-C128-C256-C512}.
 After the last layer, a convolution is applied to map to a 1 dimensional output, followed by a Sigmoid function. BatchNorm is applied after every convolution layer except for the first \texttt{C64} layer. All ReLUs are leaky with slope $0.2$. \\

\noindent \textbf{Training Procedure:} 
To optimize the networks, we follow the training procedure of pix2pix~\cite{pix2pix2017}: we alternate between one gradient descent step on D, then one step on G. As suggested in \cite{gan2014}, rather than training G to minimize
$\log(1- D(x, G(x, z))$, we maximize $\log D(x, G(x, z))$. In addition, we divide the objective by 2 while optimizing D, which slows down the rate at which D learns relative to G. We use minibatch SGD and apply the Adam solver, with learning rate $0.0002$, and momentum parameters $\beta_1 = 0.5, \beta_2 = 0.999$.
All networks were trained from scratch. Weights were initialized from a Gaussian distribution with mean 0 and standard deviation 0.02. Every experiment is trained for 150 epochs, batch size 1, with random mirroring.

\subsection{Experimental Setting}
\label{sec:analysis}

We conduct experiments in $6$ different settings shown in Table~\ref{tab:exp_methods}.

\begin{table}[t]
    \centering
    \footnotesize
    \begin{tabular}{l||c|c|c|c}
        Method & Gap Distance & Histogram Loss & Weighted Bins & Averaged Statistics \\ \hline \hline
        Cond1 & \xmark & \xmark & \xmark & \xmark \\
        Cond3 & \cmark & \xmark & \xmark & \xmark \\
        HistU & \cmark & \cmark & \xmark & \xmark \\
        HistW & \cmark & \cmark & \cmark & \xmark \\
        Hist$2^\text{nd}$ & \cmark & \cmark & \cmark & \cmark
    \end{tabular}
    \caption{Summary of the $6$ experimental settings. The Cond1 is the pix2pix framework. The Cond3 is the pix2pix framework conditioned on additional space information. The HistU, HistW, and Hist$2^\text{nd}$ are our proposed functionality-aware model.}
    \label{tab:exp_methods}
\end{table}

\textbf{Cond1.}
Cond1 denotes the experiment with original pix2pix setting where only prepared jaws are inputted. Only regression and adversarial losses are used.

\textbf{Cond3.}
Cond3 denotes the experiment of the pix2pix model conditioned on additional space information. Only regression and adversarial losses are used.

\textbf{HistU.}
HistU denotes the experiment with histogram loss with uniform weighting. The hyper-parameter $\lambda_{H}$ is set to 0.001. Regression, adversarial, and functionality (histogram) losses are used.

\begin{figure}
    \centering
    \includegraphics[width=0.6\linewidth]{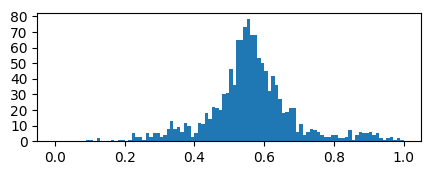}
    \caption{Distribution of the thresholds of minimal 5\% gap distances over the training set. The unit of x-axis is millimeters.}
    \label{fig:dataset_stats}
\end{figure}

\textbf{HistW.} 
To decide the weighting of histogram bins, we calculate the threshold value of minimal 5\% gap distances for each training image and display them as a histogram in Fig.~\ref{fig:dataset_stats}. The distribution of minimal 5\% gap distances peaks at around $0.5$ millimeters, which is also the critical gap distance in practice.

According to the analysis, we decide the weighting on different bins as follows. The negativity bin is weighted by $2$, the bins ranged between $0$ and $0.5$ is weighted by $1$, between $0.5$ and $1.0$ is weighted by $0.5$, and by $0$ for the rest (i.e., large gap distances). The hyper-parameter $\lambda_{H}$ is set to 0.002. Note that we do not tune the bin weighting as it is decided from the analysis. We denote this experiment as `HistW'.

\textbf{Hist$2^\text{nd}$.} Hist$2^\text{nd}$ denotes the experiment with second-order histogram loss. We use the same values for bin weighting and $\lambda_{H}$ as in `HistW'.

\subsection{Quality Assessment}
\label{sec:quality}

We evaluate the generated crowns against technicians' designs. We show that our results are comparable to the ideal designs, i.e., technicians' designs in training and validation sets, qualitatively and quantitatively.

\noindent \textbf{Metrics:} We introduce three metrics to measure quantitatively how well the predicted crowns mimic the ground truth on validation set as summarized in Table~\ref{tab:quality_val}. We do not measure the similarity on the testing set for where technicians' designs are undesirable.

\textbf{Mean root-mean-square error (RMSE): } $\text{RMSE} = \sqrt{\mathbb{E}[(\hat{y}-y)^2]}$, where $\hat{y}$ denotes the predicted crown and $y$ denotes the ground truth. RMSE is one of the standard error measurements for regression related tasks. Note that we only measure the errors in the crowns' regions to accurately assess the per-pixel quality.

\textbf{Mean Intersection-over-Union (IOU): } $\text{IOU} = (\hat{y} \cap y) / (\hat{y} \cup y)$. IOU is widely used to measure the region overlap in segmentation. We use IOU to determine whether crowns can possibly fit well into the dentition.

\textbf{Precision, Recall, and F-Measure of contours: } While mean IOU measures the quality of large areas of predicted crowns, we also introduce the boundary measurement commonly used in the contour detection task~\cite{arbelaez2011contour} to accurately assess the boundary quality of predicted crowns.

Table~\ref{tab:quality_val} shows that all the methods achieve comparable and satisfying results. The mean RMSE falls under $0.07$, the mean IOU is above $91\%$, and all the boundary-related metrics are above $93\%$. The results show that while statistical features are introduced to incorporate the constraints, they hardly sacrifice the quality of predictions.

Fig.~\ref{fig:3D} shows sample prediction results, visualized in 3D.\footnote{2D results can be found in the supplementary material.} The visualization shows that our generated crowns have similar appearances and fitting compared to ground truth. However, our methods produce more complicated surfaces for biting and chewing. For example, in cases \#1, \#3, and \#4, the generated crowns by Hist$2^\text{nd}$ have more ridges on the side that can potentially touch the opposing teeth while chewing. On the other hand, without space information, the generated crowns often over-grow as in cases \#2 and \#5.

\begin{table}[]
    \centering
    \begin{tabular}{l|| c | c || c | c | c}
         Method & Mean RMSE $\downarrow$ & Mean IOU $\uparrow$ & Precision $\uparrow$ & Recall $\uparrow$ & F-Measure $\uparrow$ \\ \hline
         Cond1 & 0.078 & 0.915 & 0.932 & 0.944 & 0.938 \\ 
         Cond3 & 0.066 & 0.922 & 0.944 & 0.953 & 0.949 \\ 
         HistU & 0.065 & 0.921 & 0.937 & 0.952 & 0.945 \\ 
         HistW & 0.066 & 0.920 & 0.931 & 0.954 & 0.942 \\ 
         Hist$2^\text{nd}$ & 0.069 & 0.916 & 0.930 & 0.947 & 0.938 \\ 
    \end{tabular}
    \caption{Quality evaluation on the validation set.}
    \label{tab:quality_val}
\end{table}

\begin{figure}
    \centering
    \includegraphics[width=1.0\textwidth]{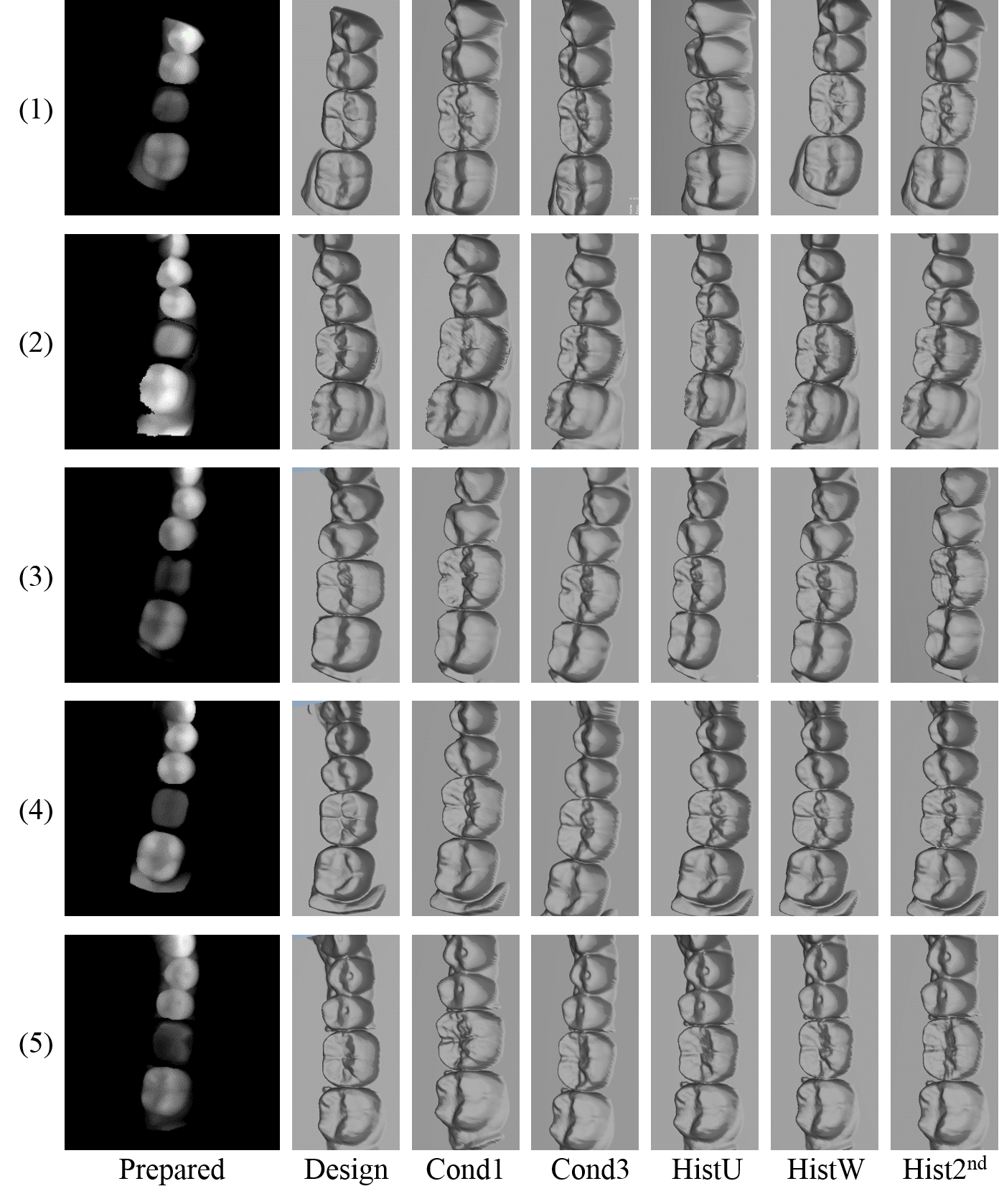}
    \caption{We visualize all the reconstructed crows in 3D model with neighboring teeth on validation set. It shows that our generated crowns have similar appearances and fitting. However, our methods produce more complicated surfaces for biting and chewing compared to technicians' designs. (3D rendering viewpoints might differ for different instances.)}
    \label{fig:3D}
\end{figure}

\subsection{Penetration Evaluation}
\label{sec:penetration}

\begin{table}[b]
    \centering
    \begin{tabular}{l||c | c || c | c || c | c}
         Method & PR(\%) val $\downarrow$ & PR(\%) test $\downarrow$ & MP val $\downarrow$ & MP test $\downarrow$ & PA val $\downarrow$ & PA test $\downarrow$ \\ \hline
         Design & - & 100.00 & - & 5.63 & - & 57.07 \\ 
         Cond1 & 53.25 & 85.60 & 12.40 & 16.47 & 101.06 & 186.39 \\ 
         Cond3 & 1.66 & 17.28 & 4.50 & 3.36 & 14.54 & 16.48 \\ 
         HistU & 1.02 & 13.99 & 4.56 & 2.62 & 15.56 & 12.24 \\ 
         HistW & 0.96 & 9.47 & 3.67 & \textcolor{red}{\textbf{2.30}} & 16.87 & \textcolor{red}{\textbf{10.00}} \\ 
         Hist$2^\text{nd}$ & 0.96 & \textcolor{red}{\textbf{7.82}} & 5.00 & 2.47 & 24.74 & 20.27 \\ 
    \end{tabular}
    \caption{Penetration evaluation results. `val' denotes validation set and `test' denotes testing set. PR denotes penetration rate. MP denotes maximum penetration. PA denotes penetration area.}
    \label{tab:penetration}
\end{table}

We evaluate the penetration with different methods on validation and testing sets. Once the crown penetrates into the opposing teeth, it is inadmissible and requires human intervention. We thus require the method produces as minimal penetration as possible.

We use the following metrics to evaluate the severity of penetration: 

\textbf{Penetration rate:} $\mathbb{E}[1(\min f(\cdot) < 0)]$, where $1(\cdot)$ is the indicator function and $f(\cdot)$ denotes the reconstructed gap distances. That is, we calculate the ratio of number of failure cases (where $\min f(\cdot) < 0$) to the total number of cases. The penetration rate is the most important metric as to measure the production failure rate.

\textbf{Average maximum penetration: } $\mathbb{E}[\min f(\cdot) \text{ if } \min f(\cdot)<0]$. That is, for each failure case, we calculate the maximum penetration value and average over all failure cases.

\textbf{Average penetration areas}: $\mathbb{E}[\#\{\min f(\cdot)<0\} \text{ if } \min f(\cdot)<0]$. That is, we measure the areas where the crown penetrates the opposing teeth and average over all failure cases.
 
\begin{figure}[t]
    \centering
    \includegraphics[width=1.0\textwidth]{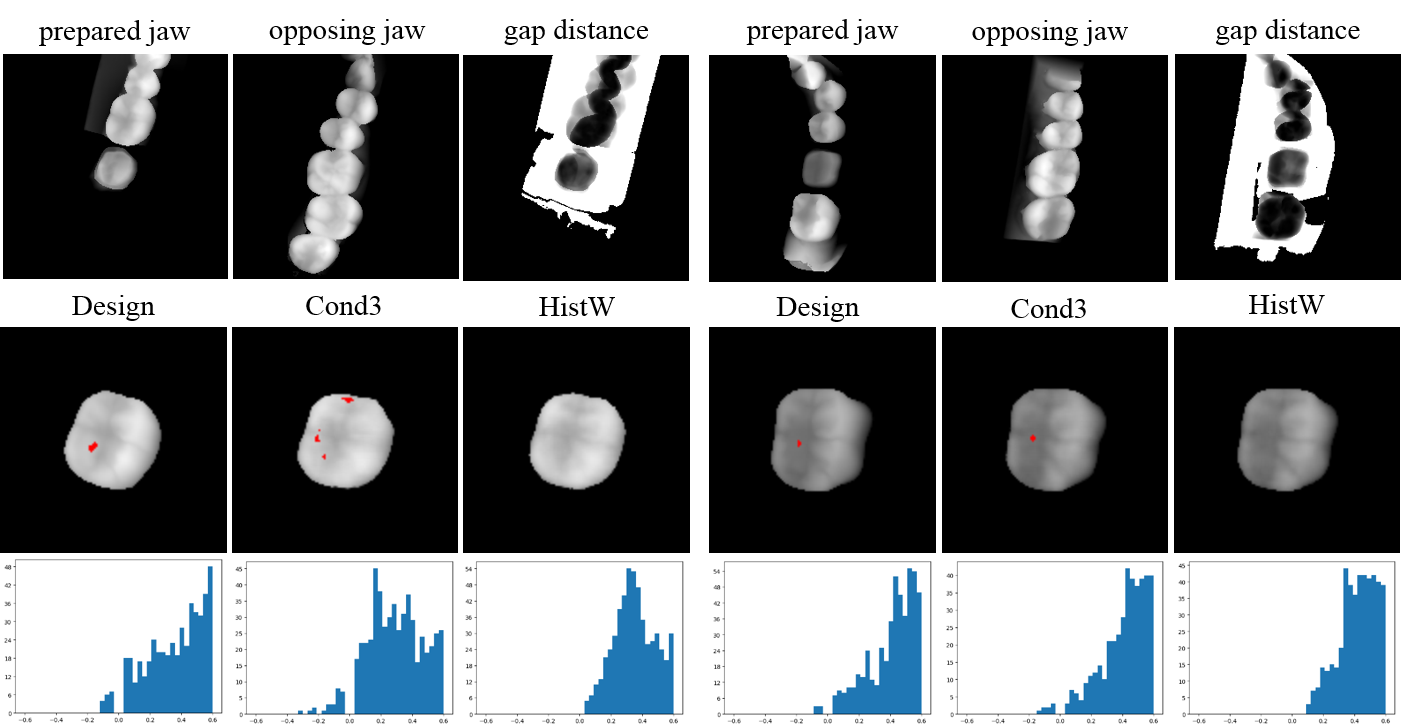}
    \caption{Visualization of violation cases on testing set. The red pixels on the crowns indicate penetration (with negative gap distances). The histogram of reconstructed gap distances is shown in the critical range between $-0.5$ and $0.5$ millimeters.}
    \label{fig:violation}
\end{figure}

We summarize the penetration evaluation in Table~\ref{tab:penetration} on both validation and testing sets. While Cond3 conditioned on space information largely prevents penetration as opposed to the Cond1, HistW and Hist$2^\text{nd}$ further reduce the penetration rate compared to the Cond3 relatively by $42\%$ on validation set and $55\%$ on testing set consistently. The average maximum penetration and penetration areas are also reduced in HistW. Hist$2^\text{nd}$, however, has larger average penetration areas. As it encourages smoothness of local regions, region-wise penetration occurs more frequently. Visualization is shown in Fig.~\ref{fig:violation}.

\subsection{Contact Point Analysis} 
\label{sec:biting}

We evaluate the contact points with different methods on validation and testing sets. As explained in subsection~\ref{sec:analysis}, we classify the crown regions within $5\%$ minimal gap distance as critical pixels that are used for biting and chewing.

\begin{table}[b]
    \centering
    \begin{tabular}{l|| c | c || c | c|| c | c || c |c }\footnotesize
         Method & NC val & Dv(\%) & NC test & Dv(\%) & Spd val & Dv(\%)& Spd test & Dv(\%)\\ \hline
         Ideal & 4.15 & 0.00 & 4.15 & 0.00 & 11.01 & 0.00 & 11.01 & 0.00 \\ 
         Design & - & - & 1.98 & -52.29 & - & - & 8.67 & -21.25 \\
         Cond1 & 1.78 & -57.11 & 1.76 & -57.59 & 12.47 & 13.26 & 14.84 & 34.79 \\ 
         Cond3 & 3.61 & -13.01 & 2.94 & -29.16 & 10.98 & -0.27 & 9.38 & -14.80 \\ 
         HistU & 3.72 & -10.36 & 3.02 & -27.23  & 11.15 & 1.27 & 9.45 & -14.17 \\ 
         HistW & 3.82 & -7.95 & 3.14 & -24.34 & 11.06 & 0.45 & 9.63 & -12.53 \\ 
         Hist$2^\text{nd}$ & 3.79 & -8.67 & \textcolor{red}{\textbf{3.74}} & \textcolor{red}{\textbf{-9.88}} & 11.05 & 0.36 & \textcolor{red}{\textbf{11.08}} & \textcolor{red}{\textbf{0.64}} \\ 
    \end{tabular}
    \caption{Contact point evaluation results. `val' denotes validation set and `test' denotes testing set. NC denotes number of clusters. Spd denotes spread. Dv denotes the deviation of the corresponding measurement to its left.}
    \label{tab:biting}
\end{table}

We use the following metrics to evaluate the distribution of contact points: 

\textbf{Average number of clusters: } We connect neighboring critical pixels within distance $2$ to form a cluster. Such a cluster forms a biting anchor. Human teeth naturally consist of several biting anchors for one tooth. Therefore, the number of clusters reflects the biting and chewing quality of generated crowns.

\textbf{Average Spread: } We measure the spatial standard deviation of critical pixels in a crown as spread. The contact points of natural teeth should scatter rather than focus on one region.

\textbf{ Deviation: } Since the number of clusters and spread have no clear trend, we measure the deviation for both number of clusters and spread. We calculate the relative error, i.e., $\frac{\hat{y}-y}{y}$, of each method as opposed to the ideal value. The ideal value is calculated using the training set which is deemed good technicians' designs.

\begin{figure}[t]
    \centering
    \includegraphics[width=1.0\textwidth]{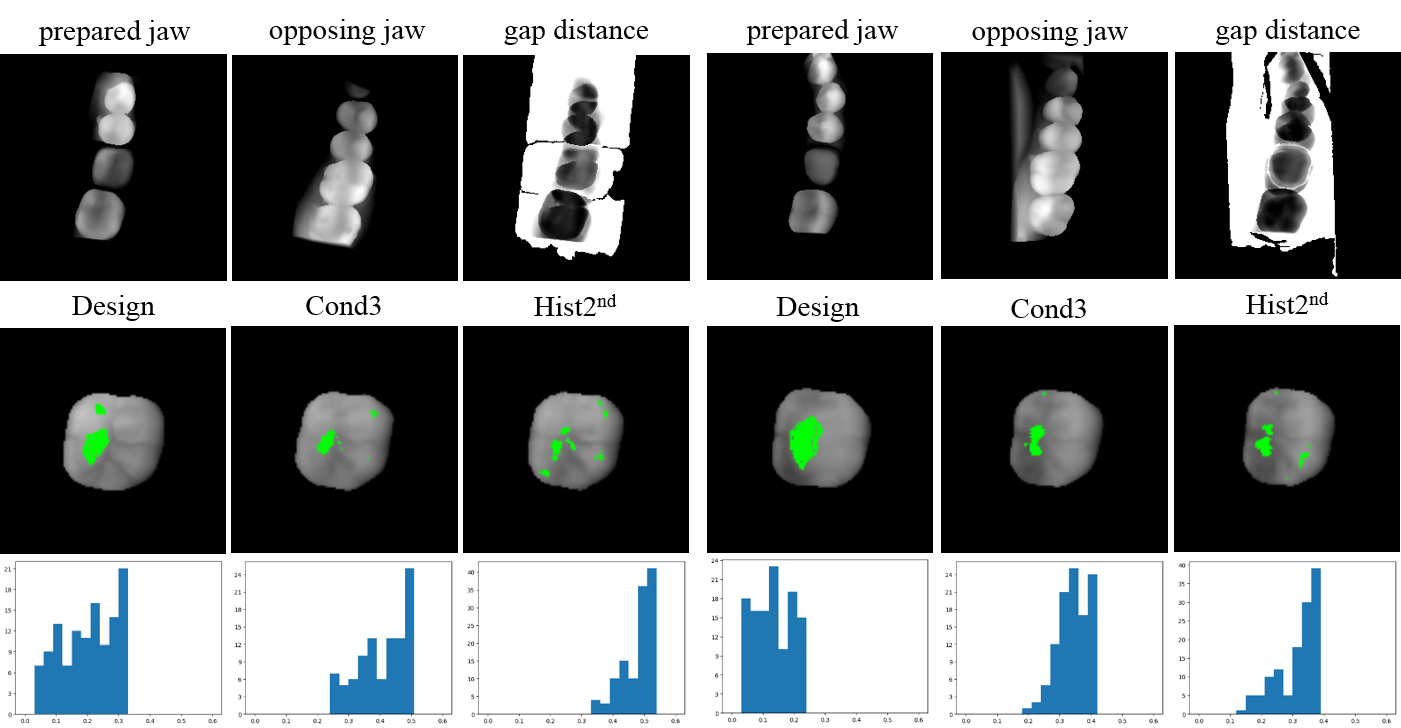}
    \caption{Visualization for contact points on testing set. The green pixels on the crowns indicate the critical regions (within 5\% minimal gap distances) that can be used to bite and chew. The histogram of 5\% minimal reconstructed gap distances is shown in the bottom row.}
    \label{fig:biting}
\end{figure}

We summarize the contact point evaluation in Table~\ref{tab:biting} on both validation and testing sets. A similar trend as in the penetration evaluation is observed: Cond3 conditioned on space information largely improves the contact points as opposed to Cond1. HistU, HistW, and Hist$2^\text{nd}$ each improves more from Cond3. The best performing Hist$2^\text{nd}$ improves from Cond3 relatively by 27.2\% and 18.1\% of number of clusters and spread respectively on testing set. It shows that while the Hist$2^\text{nd}$ performs comparably as HistW on penetration, it generally produces crowns with better contact point distributions, resulting in better biting and chewing functionality. Visualization is shown in Fig.~\ref{fig:biting} and the observations are consistent to the quantitative results.

\section{Conclusion}

We present an approach to automate the designs of dental crowns using generative models. The generated crowns not only reach similar morphology quality as human experts' designs but support better functionality enabled by learning through statistical features. This work is one of the first successful approaches to use GANs to solve a real existing problem.

\bibliographystyle{splncs}
\bibliography{dental}

\begin{thebibliography}{10}

\bibitem{krizhevsky2012imagenet}
Krizhevsky, A., Sutskever, I., Hinton, G.E.:
\newblock Imagenet classification with deep convolutional neural networks.
\newblock In: NIPS. (2012)

\bibitem{he2016deep}
He, K., Zhang, X., Ren, S., Sun, J.:
\newblock Deep residual learning for image recognition.
\newblock In: CVPR. (2016)

\bibitem{summitt2006fundamentals}
Summitt, J.B., Robbins, J.W., Hilton, T.J., Schwartz, R.S., dos Santos~Jr, J.:
\newblock Fundamentals of operative dentistry: a contemporary approach.
\newblock Quintessence Pub. (2006)

\bibitem{padbury2003interactions}
Padbury, A., Eber, R., Wang, H.L.:
\newblock Interactions between the gingiva and the margin of restorations.
\newblock Journal of clinical periodontology \textbf{30}(5) (2003)  379--385

\bibitem{zheng2011novel}
Zheng, S.X., Li, J., Sun, Q.F.:
\newblock A novel 3d morphing approach for tooth occlusal surface
  reconstruction.
\newblock Computer-Aided Design \textbf{43}(3) (2011)  293--302

\bibitem{pix2pix2017}
Isola, P., Zhu, J.Y., Zhou, T., Efros, A.A.:
\newblock Image-to-image translation with conditional adversarial networks.
\newblock In: CVPR. (2017)

\bibitem{groten2000determination}
Groten, M., Axmann, D., Pr{\"o}bster, L., Weber, H.:
\newblock Determination of the minimum number of marginal gap measurements
  required for practical in vitro testing.
\newblock Journal of Prosthetic Dentistry \textbf{83}(1) (2000)  40--49

\bibitem{gan2014}
Goodfellow, I., Pouget-Abadie, J., Mirza, M., Xu, B., Warde-Farley, D., Ozair,
  S., Courville, A., Bengio, Y.:
\newblock Generative adversarial nets.
\newblock In: NIPS. (2014)

\bibitem{cgan2014}
Mirza, M., Osindero, S.:
\newblock Conditional generative adversarial nets.
\newblock arXiv preprint arXiv:1411.1784 (2014)

\bibitem{context2016}
Pathak, D., Kr\"ahenb\"uhl, P., Donahue, J., Darrell, T., Efros, A.:
\newblock Context encoders: Feature learning by inpainting.
\newblock (2016)

\bibitem{smolensky1986information}
Smolensky, P.:
\newblock Information processing in dynamical systems: Foundations of harmony
  theory.
\newblock Technical report, COLORADO UNIV AT BOULDER DEPT OF COMPUTER SCIENCE
  (1986)

\bibitem{hinton2006reducing}
Hinton, G.E., Salakhutdinov, R.R.:
\newblock Reducing the dimensionality of data with neural networks.
\newblock Science \textbf{313}(5786) (2006)  504--507

\bibitem{vincent2008extracting}
Vincent, P., Larochelle, H., Bengio, Y., Manzagol, P.A.:
\newblock Extracting and composing robust features with denoising autoencoders.
\newblock In: ICML. (2008)

\bibitem{efros1999texture}
Efros, A.A., Leung, T.K.:
\newblock Texture synthesis by non-parametric sampling.
\newblock In: ICCV. (1999)

\bibitem{kingma2013auto}
Kingma, D.P., Welling, M.:
\newblock Auto-encoding variational bayes.
\newblock In: ICLR. (2014)

\bibitem{oord2016pixel}
Oord, A.v.d., Kalchbrenner, N., Kavukcuoglu, K.:
\newblock Pixel recurrent neural networks.
\newblock PMLR (2016)

\bibitem{van2016conditional}
van~den Oord, A., Kalchbrenner, N., Espeholt, L., Vinyals, O., Graves, A.,
  et~al.:
\newblock Conditional image generation with pixelcnn decoders.
\newblock In: NIPS. (2016)

\bibitem{arjovsky2017towards}
Arjovsky, M., Bottou, L.:
\newblock Towards principled methods for training generative adversarial
  networks.
\newblock In: ICLR. (2017)

\bibitem{chen2016infogan}
Chen, X., Duan, Y., Houthooft, R., Schulman, J., Sutskever, I., Abbeel, P.:
\newblock Infogan: Interpretable representation learning by information
  maximizing generative adversarial nets.
\newblock In: NIPS. (2016)

\bibitem{denton2015deep}
Denton, E.L., Chintala, S., Fergus, R.,  et~al.:
\newblock Deep generative image models using a laplacian pyramid of adversarial
  networks.
\newblock In: NIPS. (2015)

\bibitem{donahue2016adversarial}
Donahue, J., Kr{\"a}henb{\"u}hl, P., Darrell, T.:
\newblock Adversarial feature learning.
\newblock In: ICLR. (2016)

\bibitem{dumoulin2016adversarially}
Dumoulin, V., Belghazi, I., Poole, B., Mastropietro, O., Lamb, A., Arjovsky,
  M., Courville, A.:
\newblock Adversarially learned inference.
\newblock In: ICLR. (2016)

\bibitem{radford2015unsupervised}
Radford, A., Metz, L., Chintala, S.:
\newblock Unsupervised representation learning with deep convolutional
  generative adversarial networks.
\newblock In: ICLR. (2016)

\bibitem{reed2016generative}
Reed, S., Akata, Z., Yan, X., Logeswaran, L., Schiele, B., Lee, H.:
\newblock Generative adversarial text to image synthesis.
\newblock (2016)

\bibitem{zhang2017stackgan}
Zhang, H., Xu, T., Li, H., Zhang, S., Huang, X., Wang, X., Metaxas, D.:
\newblock Stackgan: Text to photo-realistic image synthesis with stacked
  generative adversarial networks.
\newblock In: ICCV. (2017)

\bibitem{zhao2016energy}
Zhao, J., Mathieu, M., LeCun, Y.:
\newblock Energy-based generative adversarial network.
\newblock In: ICLR. (2017)

\bibitem{zhu2016generative}
Zhu, J.Y., Kr{\"a}henb{\"u}hl, P., Shechtman, E., Efros, A.A.:
\newblock Generative visual manipulation on the natural image manifold.
\newblock In: ECCV. (2016)

\bibitem{karras2017progressive}
Karras, T., Aila, T., Laine, S., Lehtinen, J.:
\newblock Progressive growing of gans for improved quality, stability, and
  variation.
\newblock ICLR (2018)

\bibitem{sohn2015learning}
Sohn, K., Lee, H., Yan, X.:
\newblock Learning structured output representation using deep conditional
  generative models.
\newblock In: NIPS. (2015)

\bibitem{guadarrama2017pixcolor}
Guadarrama, S., Dahl, R., Bieber, D., Norouzi, M., Shlens, J., Murphy, K.:
\newblock Pixcolor: Pixel recursive colorization.
\newblock In: BMVC. (2017)

\bibitem{walker2016uncertain}
Walker, J., Doersch, C., Gupta, A., Hebert, M.:
\newblock An uncertain future: Forecasting from static images using variational
  autoencoders.
\newblock In: ECCV. (2016)

\bibitem{xue2016visual}
Xue, T., Wu, J., Bouman, K., Freeman, B.:
\newblock Visual dynamics: Probabilistic future frame synthesis via cross
  convolutional networks.
\newblock In: NIPS. (2016)

\bibitem{wang2016generative}
Wang, X., Gupta, A.:
\newblock Generative image modeling using style and structure adversarial
  networks.
\newblock In: ECCV. (2016)

\bibitem{mathieu2015deep}
Mathieu, M., Couprie, C., LeCun, Y.:
\newblock Deep multi-scale video prediction beyond mean square error.
\newblock In: ICLR. (2016)

\bibitem{yoo2016pixel}
Yoo, D., Kim, N., Park, S., Paek, A.S., Kweon, I.S.:
\newblock Pixel-level domain transfer.
\newblock In: ECCV. (2016)

\bibitem{ledig2016photo}
Ledig, C., Theis, L., Husz{\'a}r, F., Caballero, J., Cunningham, A., Acosta,
  A., Aitken, A., Tejani, A., Totz, J., Wang, Z.,  et~al.:
\newblock Photo-realistic single image super-resolution using a generative
  adversarial network.
\newblock In: CVPR. (2017)

\bibitem{zhu2017unpaired}
Zhu, J.Y., Park, T., Isola, P., Efros, A.A.:
\newblock Unpaired image-to-image translation using cycle-consistent
  adversarial networks.
\newblock In: ICCV. (2017)

\bibitem{zhu2017toward}
Zhu, J.Y., Zhang, R., Pathak, D., Darrell, T., Efros, A.A., Wang, O.,
  Shechtman, E.:
\newblock Toward multimodal image-to-image translation.
\newblock In: NIPS. (2017)

\bibitem{lazebnik2006beyond}
Lazebnik, S., Schmid, C., Ponce, J.:
\newblock Beyond bags of features: Spatial pyramid matching for recognizing
  natural scene categories.
\newblock In: CVPR. (2006)

\bibitem{perronnin2010improving}
Perronnin, F., S{\'a}nchez, J., Mensink, T.:
\newblock Improving the fisher kernel for large-scale image classification.
\newblock In: ECCV. (2010)

\bibitem{carreira2012semantic}
Carreira, J., Caseiro, R., Batista, J., Sminchisescu, C.:
\newblock Semantic segmentation with second-order pooling.
\newblock In: ECCV. (2012)

\bibitem{simonyan2013deep}
Simonyan, K., Vedaldi, A., Zisserman, A.:
\newblock Deep fisher networks for large-scale image classification.
\newblock In: NIPS. (2013)

\bibitem{gong2014multi}
Gong, Y., Wang, L., Guo, R., Lazebnik, S.:
\newblock Multi-scale orderless pooling of deep convolutional activation
  features.
\newblock In: ECCV. (2014)

\bibitem{wang2016learnable}
Wang, Z., Li, H., Ouyang, W., Wang, X.:
\newblock Learnable histogram: Statistical context features for deep neural
  networks.
\newblock In: ECCV. (2016)

\bibitem{ustinova2016learning}
Ustinova, E., Lempitsky, V.:
\newblock Learning deep embeddings with histogram loss.
\newblock In: NIPS. (2016)

\bibitem{unet2015}
Ronneberger, O., Fischer, P., Brox, T.:
\newblock U-net: Convolutional networks for biomedical image segmentation.
\newblock In: International Conference on Medical image computing and
  computer-assisted intervention, Springer (2015)  234--241

\bibitem{arbelaez2011contour}
Arbelaez, P., Maire, M., Fowlkes, C., Malik, J.:
\newblock Contour detection and hierarchical image segmentation.
\newblock PAMI (2011)

\end{thebibliography}

\section{Appendix}
\subsection{2D Results in Fig. 5 (Main Text)}

\begin{figure}
    \centering
    \includegraphics[width=\textwidth]{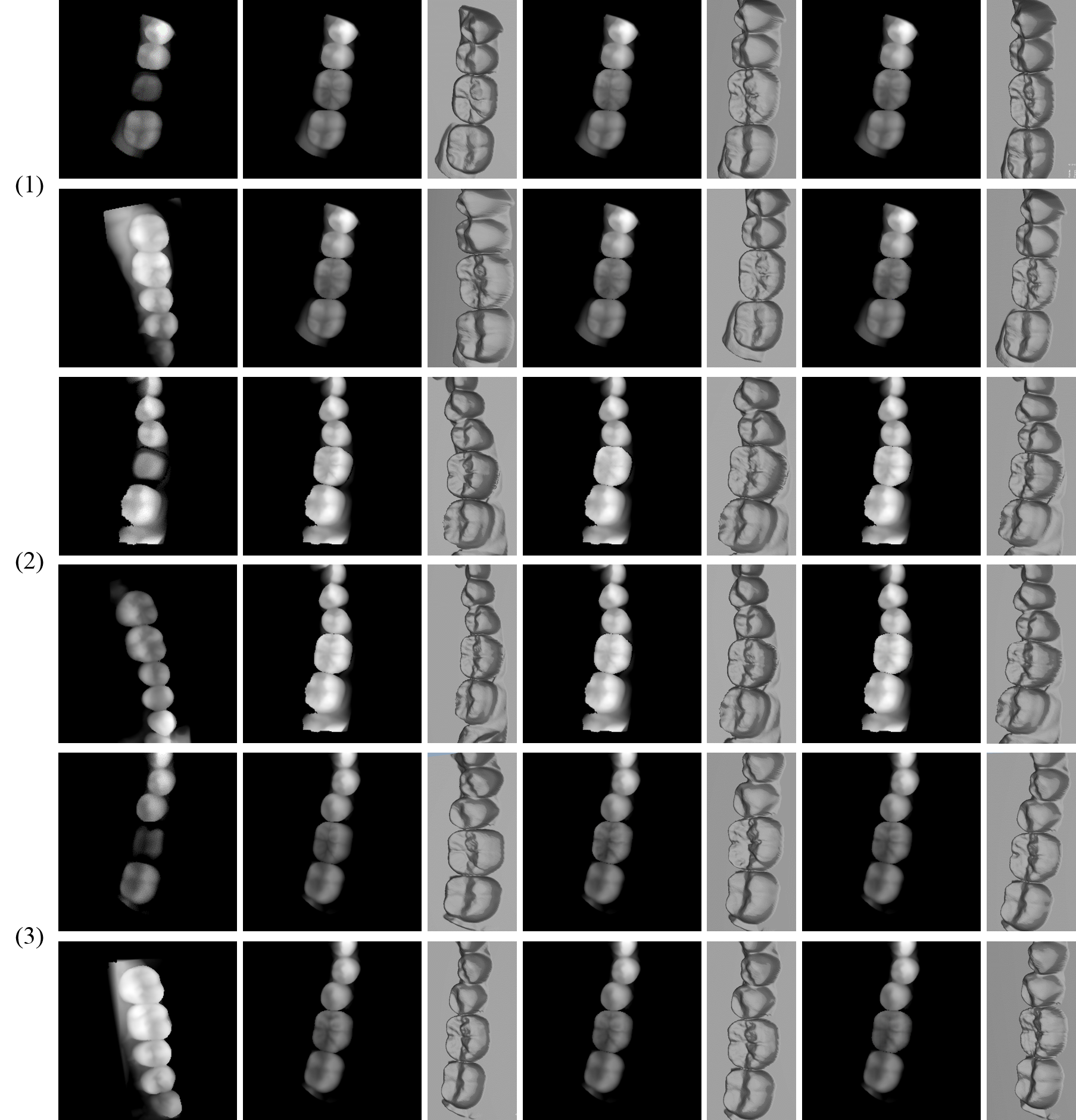}
    \caption{The case number corresponds to the same case in the main text and each case has two rows. Left to right in the first row: prepared jaw, design, cond1, cond3. Left to right in the second row: opposing jaw, HistU, HistW, Hist$2^\text{nd}$.}
    \label{fig:results1}
\end{figure}

\begin{figure}
    \centering
    \includegraphics[width=\textwidth]{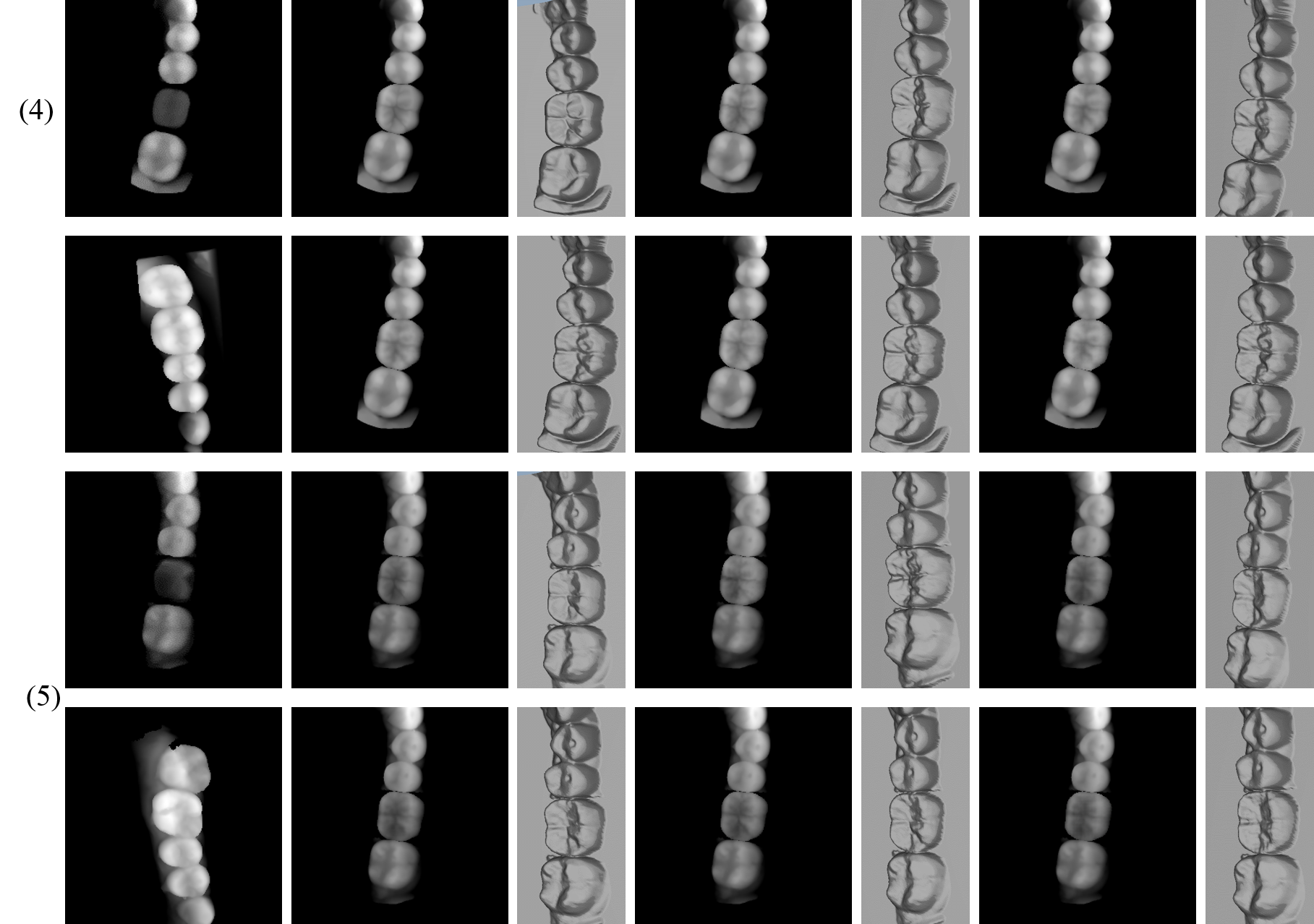}
    \caption{The case number corresponds to the same case in the main text and each case has two rows. Left to right in the first row: prepared jaw, design, cond1, cond3. Left to right in the second row: opposing jaw, HistU, HistW, Hist$2^\text{nd}$.}
    \label{fig:results2}
\end{figure}

\end{document}